# Exploring the role of Input and Output Layers of a Deep Neural Network in Adversarial Defense

Jay N. Paranjape
Computer Science & Engineering
Indian Institute of Technology
Delhi, India
email: jay.edutech@gmail.com

Rahul Kumar Dubey
Robert Bosch Engineering and
Business Solutions
Bangalore, India
e-mail:
RahulKumar.Dubey@in.bosch.com

Vijendran V. Gopalan
Robert Bosch Engineering and
Business Solutions
Bangalore, India
e-mail:
GopalanVijendran.Venkaparao@
in.bosch.com

*Abstract* - **Deep neural networks are learning models having achieved state of the art performance in many fields like prediction, computer vision, language processing and so on. However, it has been shown that certain inputs exist which would not trick a human normally, but may mislead the model completely. These inputs are known as adversarial inputs. These inputs pose a high security threat when such models are used in real world applications. In this work, we have analyzed the resistance of three different classes of fully connected dense networks against the rarely tested non-gradient based adversarial attacks. These classes are created by manipulating the input and output layers. We have proven empirically that owing to certain characteristics of the network, they provide a high robustness against these attacks, and can be used in fine tuning other models to increase defense against adversarial attacks.**

## I. INTRODUCTION

Today, many real world applications like Automated Driving[1], Anomaly Detection[2] and applications in Healthcare[3], are being realized using Deep Learning algorithms. This is all the more reason to focus on security when using these algorithms. Szegedy et al. first showed that it is possible to generate inputs, which a human is able to classify correctly but which completely mislead state-of-the-art models, by only adding very small perturbations to the original image [4]. With Deep Learning algorithms being increasingly used in real world applications, a good defense mechanism against such adversarial attacks is strongly needed. There have been various attempts to increase robustness as well as to classify inputs as being normal or adversarial([5] [6] [7] [8] [9] ). However, new attacks are developed everyday which can fool the existing state-of-the-art defense mechanisms. Also, most of the works focus mainly on gradient based attacks but not many have verified their results for non gradient based attacks.

The prevalent defense techniques include changing the model architecture, including dropout or non-gradient based layers([5] [10]), or including adversarial examples at training time [11]. However, not much work has been done in exploring the effects that data manipulations have on adversarial defense. This paper explores the effects of binarizing the input image(modifying the input layer), as well as carrying out the task of image classification through image recall(modifying the output layer). We empirically observe the resistance of these changes against non gradient based attacks as implemented in Foolbox 0.10.0 [12] and showed that these changes are indeed crucial for improving the robustness for specific image classification tasks.

## II. RELATED WORK

### A. Types of adverserial attacks

There are various kinds of adversarial attacks and new attacks keep getting prepared, but on a higher level, attacks can be classified into 4 groups:

- *Gradient Based Attacks:*

These are the most prevalent attacks. They rely on detailed model information like the gradient of the loss with respect to the input. Examples include FGSM[13], C&W Attack[14] , BIM[15] and DeepFool[16]. A common defense mechanism used against gradient based attacks is masking the gradients.This blocks the information which is necessary for such attacks. Examples of defense mechanisms against these attacks include defensive distillation [5]and saturated non-linearities[17]

- *Score Based Attacks:*

These attacks, on a higher level, compute approximate gradients from the logits of the model. These include SinglePixelAttack[18] and blackbox variants of Gradient-Based attacks like C&W attack.[19]. However, numerical estimation of gradients from logits can be obstructed by introducing dropout layers in the network architecture. Defense mechanisms which hide the gradients as well as the numerical estimates were developed by Tramer et al. in 2017.[11]

- *Transfer Based Attacks:*

These attacks require information about the training data. They use the data to generate adversarial examples on an ensemble of models and rely on the fact that these examples will also be transferred to the original model, which means these will serve as adversarial examples for the original model as well. Works by Papernot et al.[20] and Liu et al[21] explore these attacks.

- *Decision Based Attacks:*

This is a recent class of attacks which require no information about the training data or gradients. They only depend upon the decision of the model for a given input. Examples include Boundary Attack[22](which succeeded in misleading many state of the art models), Additive Gaussian Noise Attack and Additive Uniform Noise Attack.[12]

Since these attacks require almost negligible information, they take longer to find an appropriate adversary for a given image. Though these attacks are computationally expensive, they could be the most relevant with respect to real world applications, where model information is generally not available.

Our work focuses mainly upon Single Pixel Attacks(Figure 4), Boundary Attacks(Figure 6) and Additive Gaussian Noise Attacks(Figure 5 and 7)

B. *Different Defense and Classification Mechanisms*

Many attempts have been made to classify adversarial inputs from normal inputs as well as generating a strong robustness towards such attacks. Some of these include the following:

- *Local Intrinsic Dimensionality(LID)*

This method was developed in 2018 by Ma et al.[9] to classify normal and noisy inputs from adversarial ones. They argued that an adversarial example should have a higher dimensionality than a normal or noisy example. They calculated the local variant of the dimensionality from maximum likelihood estimate using $k$ nearest neighbours of the particular example in the input dataset. Then they argued that the LID for an adversarial input is higher than a normal or noisy input, and based on the LID from each layer of the network, the inputs can be classified with a very high accuracy using Machine Learning Classifiers.

- *Ensemble Adversarial Training*

This method involves training on a dataset which includes adversarial examples from an ensemble of substitute models, along with the original dataset. It was developed in 2017 by Tramer et al.[11]This method has proven to be highly successful against almost all gradient based, score based as well as transfer based attacks in the 2017 Kaggle Competition on Adversarial Attacks [1]

- *Color Bit Reduction and Total Variance Minimization*

This method uses a similar approach as ours in that it focuses on image preprocessing for adversarial defense. The color bit reduction decreases the intensity according to the following formula: $X[i,j] = X[i,j] - (X[i,j] \% power(2,color\ bit))$, where color bit is a hyperparameter. [23] Total variance Minimization(TVM) makes use of denoising the image so as to reduce the variance between the pixels. We use the implementation of TVM as provided by skimage module of python for experiments[24]. Both these approaches aim to reduce the unnecessary features in images which are used by attacks.

## III PRELIMINARIES

A. *Classification versus Recall*

In general classification tasks, the network structure includes a few convolutional layers followed by dense layers, with the last layer(output layer) having nodes equal to the number of labels. However, in a recall task, the network tries to convert the input image into another image which is one of the representative images of each class. Thus, the number of output nodes is equal to the input size. The weights in the network are trained so that the input image is recalled into an image which has the least L2 distance from its corresponding representative image. An example is shown in Figure 1.

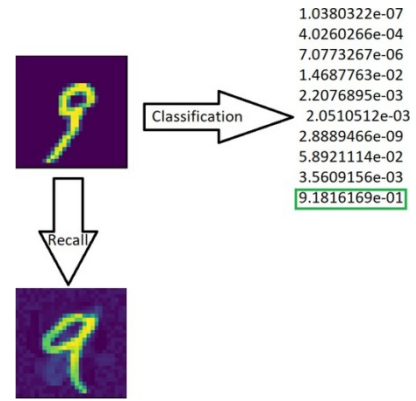

Figure 1: Classification versus Recall

B. *Notation*

In this paper, we have used the implementations of different adversarial attacks as used in Foolbox[12]. We have taken the first image of each class as the representative image for the class.

We define failed attacks as the inputs for which the attacking algorithm failed to find an attack or if found, is still classified correctly by the model. The second condition is important in case of certain restrictions on the input data like binarization.

We define Resistance as the ratio of failed attacks to the number of total inputs.

The following notation is used throughout the paper:

- $N_f$: Number of failed attacks on a dataset
- $N$ : Size of the dataset
- $\Omega$ : Resistance = $\frac{N_f}{N}$

## IV EXPERIMENTS - EFFECTS OF INPUT AND OUTPUT LAYER MODIFICATIONS

A. *Types of Models*

We have used three types of models. All of them comprise of only fully connected layers with changes made only in the input and/or output layers. These are described as below:

- IBOL ( Inputs Binarized Outputs Logits) - The inputs are binarized before passing to the model. The output nodes are equal to the number of labels. Hence, the output is a vector with size equal to the number of classes. These denote the scores or probabilities of the input belonging to a class and are known as logits.

- INOI ( Inputs Non-Binarized Outputs Images ) - the output nodes are equal to input size. Some examples are shown in figures 2 and 3.

- IBOI ( Inputs Binarized Outputs Images ) - the inputs are binarized and the outputs nodes are equal to input size. Some examples are shown in figures 2 and 3.

For INOI and IBOI, we have carried out the classification task by taking the L2 distance between the recalled image and each of the representative images and choosing the label of the representative with the lowest L2 distance. L2 distance is defined as the root of the sum of squares of the difference of pixels in the original and recalled images. In all the models, a standard 2 hidden layer neural network, with RELU activation function is used.

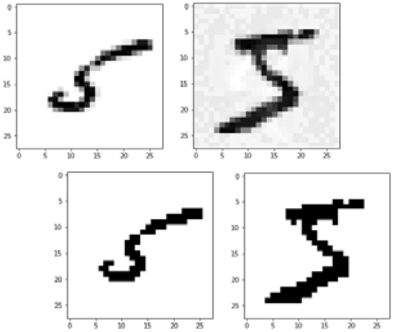

Figure 2: Digit '5' recalled to its representative image in case of INOI(top two) and IBOI(bottom two)

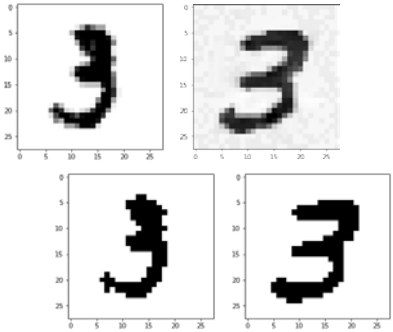

Figure 3: Digit '3' recalled to its representative image in case of INOI(top two) and IBOI(bottom two)

### B. Calculation of Resistance

Since it was computationally very expensive to compute resistance using each of the 60,000 images of the MNIST dataset, we randomly selected 500 images to act as representative of the training data in case of boundary attacks. The algorithm we followed is described as Algorithm 1:

**Algorithm 1** Calculate Ω = *Resistance*

**Require:** *model.predict(img) = labels(img)* $\forall$ *img* $\in$ *imgs*
*imgs* ← randomly select 500 images from dataset
$N_f$ ← 0
$N$ ← 500
**for all** *img* $\in$ *imgs* **do**
  *adv_image* ← an adversarial attack using foolbox

  **try:** dummy ← *adv_img.shape* {since foolbox returns none if it dosen't find adversary}
  **with:** *adv_img* ← *img*

  **if** *model.predict (adv_img) = model.predict (img)* {attack failed} **then**
    $N_f = N_f + 1$
  **end if**
**end for**
Ω ← ($N_f/N$)
**return** Ω

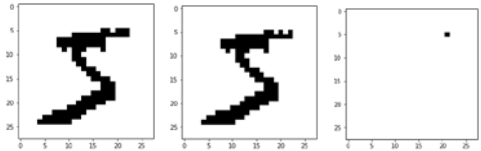

Figure 4: Example of a Single Pixel Attack on IBOL. (Left) original(5) (Center) adversary(misclassified as 3) (Right) Difference

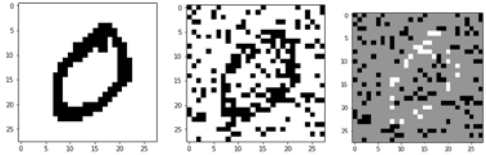

Figure 5: Example of a Additive Gaussian Attack on IBOI. (Left) original(0) (Center) adversary(2) (Right) Difference

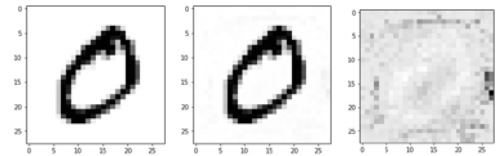

Figure 6: Example of a Boundary Attack on INOI. (Left) original(0) (Center) adversary(misclassified as 2) (Right) Difference

### V. RESULTS AND ANALYSIS

The results are tabulated in table 1. Boundary attacks are costlier in terms of time taken per image and so we selected

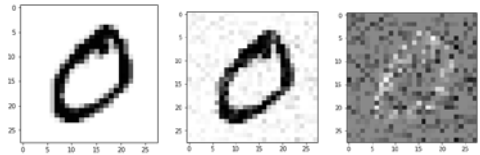

Figure 7: Example of Additive Gaussian Noise Attack on INOI. (Left) original(0) (Center) adversary(3) (Right) Difference

Table 1: Attack Resistance for Different Models

|      | SPA   | BA    | AGNA  | AUNA  |
|------|-------|-------|-------|-------|
| CDR  | 0.1   | 0.04  | 0.01  | 0.02  |
| TVM  | 0.15  | 0.05  | 0.02  | 0.02  |
| IBOL | 0.98  | 0.933 | 0.606 | 0.994 |
| INOI | 0.72  | 0.93  | 0.886 | 0.912 |
| IBOI | 0.966 | 1.0   | 0.38  | 0.994 |

SPA = Single Pixel Attack, BA = Boundary Attack, AGNA = Additive Gaussian Noise Attack, AUNA = Additive Uniform Noise Attack

500 random samples across multiple runs for calculating resistance. It was observed that there was a difference of less than 0.05 in the measured resistance across multiple measurements. Hence, we can say that the values of resistance as given in the table are quite similar to what we would have observed if the same was done for all images.

It can be seen clearly that IBOL and IBOI performs exceptionally well in all cases except for the Additive Gaussian Noise Attack. However, we don't worry about the particular case since the resulting adversary, as seen in figure 5, is quite noisy to the human eye as well. This occurs because when we binarize the image, the 'noise' reduces to either 1 or 0, which is easily distinguishable for the human eye as well. In the case of INOI where the range was 0 to 255, the noise amounts to very small changes in these values, which are not immediately perceivable for the human eye as well. We further tried other input varying layers such as Total Variance Minimization(TVM) and ColorDepth Reduction(CDR)[23] However, these techniques exhibit very poor resistance to almost all tested non gradient based attacks.

For the three different models, we can make the following observations:

1) When inputs are binarized, it highly increases resistance against decision based attacks. However, gradient based attacks are not fully eliminated since the model still has a continuous gradient. Also, binarizing an image reduces the required computational resources by a considerable amount.

2) IBOI combines properties of both (1) and (2). Hence, it exhibits high robustness against all attacks. However, due to the opposing effects of reduction in complexity by binarization and increasing the complexity through recall, this model fails to scale to larger datasets easily.

3) Boundary attack[22] has by far been the most successful. It has utmost importance in the real-world scenario, because it requires the minimum information about the model. It has deceived the previous state-of-the-art defense mechanisms like Distillation[5] with a successful attack rate of almost 100 %. As far as we know, very few papers have tested for boundary attacks. However, the IBOI type of model exhibits quite high resistance to it, though we have only verified it for the MNIST dataset.

4.) There is an unexpected drop in resistance in the case of TVM and CDR. This may be due to the fact that these techniques, as their names suggest, focus on decreasing the available feature space which can be exploited by gradient based attacks. The attacks analysed in this paper, mainly Gaussian and Uniform Noise Attacks, use a different technique(adding noise to the image) to fool models. Hence, decreasing the dimensionality of the feature space does not seem to have a desired effect on resistance

## VI. WHY DOES THIS WORK?

We review why such results are obtained and why changing the input and output styles cause increase robustness.

**Binarization** - The input image has pixels with values ranging from 0 to 255. We binarize it with 128 as the threshold. Thus, every pixel can now take a value of either 0 or 1. This drastically reduces the opportunities for 'slight' perturbations since only the pixels with values close to 128 will be affected. Hence, even after an adversarial attack, the image is still classified as the original, or it is too noisy, even for the human eye, similar to figure 5.

**Recall** - A proper representative image and appropriate training will recall an image as shown in figure 2 and 3. Even with slight perturbations to the image, the new recalled output formed will be farther away from the representative image. However, due to the increased number of attributes(depending on the size of the image), the recalled image should still be closer to the representative image of the same class than from those of other classes. In other words, the changes in the value of a single pixel are not much important in recall. Since the recalled output for an input image also looks alike to the human eye, it follows that attacks like Boundary attack, which iterate to bring the adversarial sample closer to the original image, don't work that well. This is unlike the case when the output is a single label from the culmination of all the pixels of the input image. Thus, by increasing the complexity of the output space, we try to increase robustness against Decision based attacks

## VII. DRAWBACKS

1.) Non transferability to diverse datasets - The method proposed above cannot be directly scaled to datasets like CIFAR where the images belonging to the same class are highly different from each other. This proves to be a problem in recalling to a particular image of a class. How this can be done is a topic for future work. Also, recalling an image is much more complex than the traditional classification. Hence, for classifying large and diverse datasets, major changes must be made in the architecture and training process. We verified this by running the recall model on CIFAR 10 and using ResNet but we could not get even 20 % accuracy.

2.) Choosing the representative image - The representative image is currently the first image as provided by the dataset. However, there should be a better way of choosing a representative image. It cannot simply be the image with the least error since the learning process commences after the representative selection.

3.) Binarization Technique - For binarization, we applied a simple threshold at intensity value 127. However, this can be replaced by techniques like BRINT[25] and Binarization Ensembles[26], which have been shown to be more robust to rotation and scaling, while successfully compressing the data.

## VIII. CONCLUSION

We conclude that along with the model architecture, the descriptions of the input and output layers also play a major role in deciding the robustness of the model. Also, we have shown that this method is quite resistant against the recently developed, effective and the least tested decision attacks like the boundary attack. We observed that recalling an image is much more complicated than outputting logits but is highly resistant to adversarial attacks. Such attacks target the high dimensionality of the input, and so there arises a tradeoff between accuracy and robustness due to compression of the image. We think that finding and improving techniques to involve compression without harming the overall classification accuracy is an interesting problem.

We believe that this architecture can replace the dense layers at the head of existing state-of-the-art models and can increase robustness. The proposed method, as of now has drawbacks for image recognition. However, when images of a class are highly similar to each other as in MNIST dataset, we expect that fine tuning with IBOI should greatly render robustness to the overall model. This can find potential applications in the robust Non Destructive Evaluation of machine parts in industry.

## IX. ACKNOWLEDGEMENTS

The authors gratefully acknowledge Robert Bosch for the opportunity to intern and the Indian Institute of Technology Delhi for their permission.